\documentclass{article}

\usepackage{microtype}
\usepackage{graphicx}
\usepackage{subcaption}
\usepackage{booktabs} 

\usepackage{hyperref}

\usepackage{hyperref}       %
\usepackage{url}            %
\usepackage{booktabs}       %
\usepackage{amsfonts}       %
\usepackage{nicefrac}       %
\usepackage{microtype}      %
\usepackage[table]{xcolor}  %
\usepackage{amsmath}
\usepackage{enumitem}

\usepackage{graphicx}
\usepackage{caption}
\usepackage{cleveref}
\usepackage{multirow}
\usepackage{cuted}

\usepackage{subcaption}
\usepackage{algorithm}   

\usepackage{wrapfig}
\usepackage{adjustbox}
 
\usepackage{listings} 
\usepackage[most]{tcolorbox}
\usepackage{tikz}

\usepackage{subcaption}
 
\usepackage{booktabs}
\usepackage{tabularx}

\usepackage[preprint]{icml2026}

\usepackage{amsmath}
\usepackage{amssymb}
\usepackage{mathtools}
\usepackage{amsthm}

\theoremstyle{plain}

\theoremstyle{definition}

\theoremstyle{remark}

\usepackage[textsize=tiny]{todonotes}

\icmltitlerunning{Query Your Way Out: Benchmarking Multi-Turn Knowledge Reasoning through Generalized Puzzle Solving}

\usepackage{tcolorbox}

\newtcolorbox{mytextbox}[1][]{
    colback=white,
    colframe=black,
    boxrule=2pt,
    arc=3pt,
    left=8pt,
    right=8pt,
    top=8pt,
    bottom=8pt,
    fonttitle=\bfseries,
    title={#1}
}
\renewcommand{\today}{October 11, 2026}
\begin{document}

\twocolumn[
  \icmltitle{Query Your Way Out: Benchmarking Multi-Turn Knowledge \\Reasoning through Generalized Puzzle Solving}



  \icmlsetsymbol{equal}{*}

    \begin{icmlauthorlist}
      \icmlauthor{Xin Cheng}{seu}
      \icmlauthor{Junming Yang}{seu}
      \icmlauthor{Ning Xu}{seu}
    \end{icmlauthorlist}

    \icmlaffiliation{seu}{School of Computer Science and Engineering, Southeast University, Nanjing, China}

    \icmlcorrespondingauthor{Ning Xu}{xning@seu.edu.cn}

  \icmlkeywords{Machine Learning, ICML}

  \vskip 0.3in
]


\printAffiliationsAndNotice{}  

\begin{abstract}
Puzzles offer a natural testbed for Large Language Models (LLMs) reasoning because they pose verifiable goals under partial information and require multi-step decisions to reach a solution. However, existing puzzle benchmarks typically focus on narrow environments with rigid mechanics and tightly constrained solution targets. To bridge this gap, we introduce the Active Probing Benchmark (APB), which generalizes puzzle-solving by casting concept identification as an interactive reasoning task that integrates specialized knowledge assessment into the puzzle setting. Specifically, APB evaluates whether models can acquire information via multi-turn active probing and inference to pinpoint 960 target concepts of specialized difficulty across eight diverse academic domains. Our evaluation reveals a counter-intuitive phenomenon: most open-source reasoning models significantly underperform their standard instruction counterparts. This finding suggests that in interactive settings, these models struggle to leverage procedural reasoning to reliably access and apply specialized domain knowledge. Beyond these empirical insights, APB provides a scalable pipeline for generating benchmarks that probe knowledge reasoning under interaction across diverse settings.
\end{abstract}
\section{Introduction}

\begin{figure*}[t!]
\centering
\includegraphics[width=0.92\textwidth]{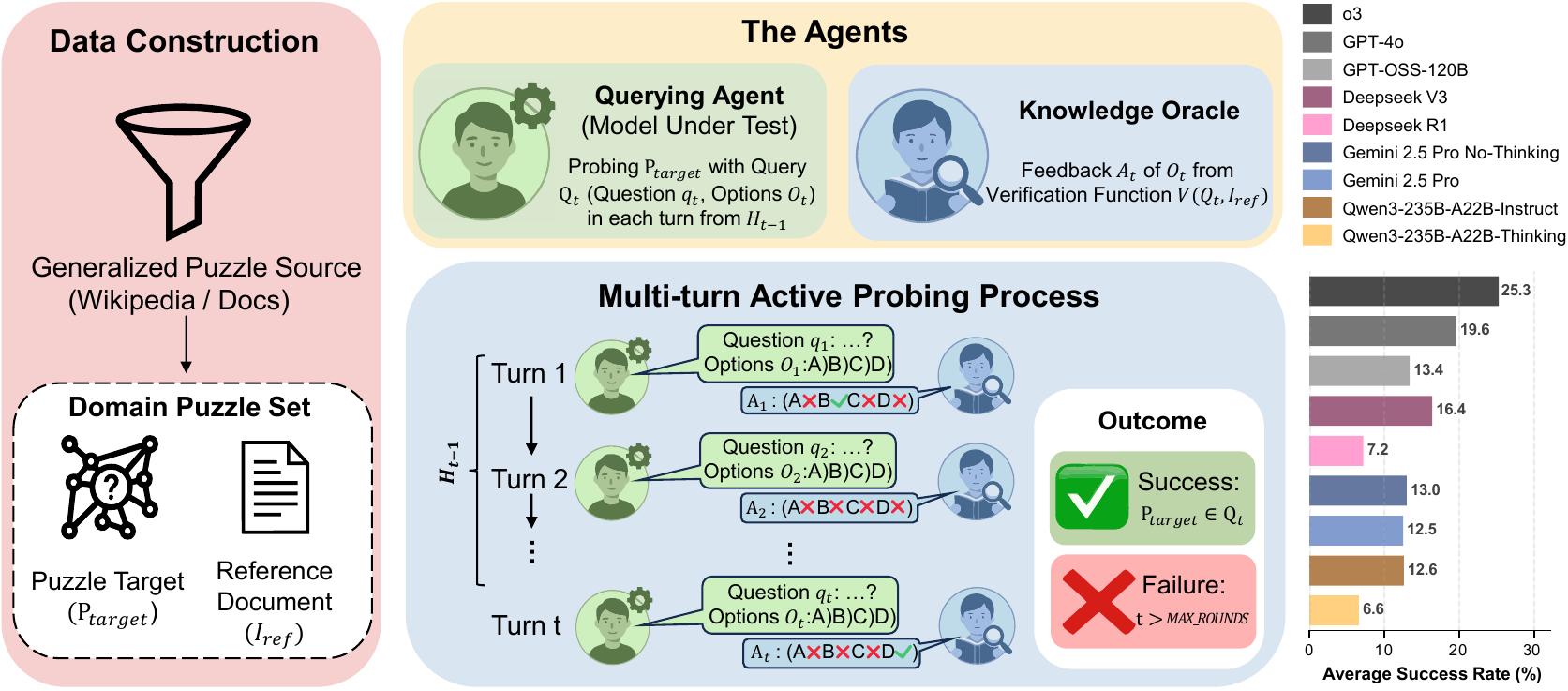}
\caption{Overview of the Active Probing Benchmark (APB) and representative performance results. \textbf{(Left)} \textit{Data construction}: each puzzle specifies a target concept $P_{\text{target}}$ drawn from a generalized knowledge source (e.g., Wikipedia or domain documents) along with a reference document $I_{\text{ref}}$. \textbf{(Middle)} \textit{Multi turn active probing}: the model under test iteratively asks multiple-choice queries $Q_t$, receives feedback $A_t$ from a knowledge oracle, and updates its hypotheses to reduce the candidate space until the target is identified or a turn budget is exhausted. \textbf{(Right)} \textit{Performance comparison}: average success rates show that several open source ``Thinking'' models underperform their ``No-Thinking'' counterparts. Full results are reported in Table~\ref{tab:model_performance}.}
\label{fig:framework}
\end{figure*}

The rapid progress of Large Language Models (LLMs) has heightened the demand for evaluation paradigms that reliably reflect underlying model capabilities~\citep{singh2024evaluation}. Although existing benchmarks have enabled systematic comparison, their validity is increasingly constrained by practical challenges in benchmark design and upkeep. Prior work has noted that overlap between evaluation data and large-scale pre-training corpora may confound performance estimates in some settings~\citep{li2023open,jacovi2023stop,roberts2023cutoff}, motivating interest in evaluation protocols that emphasize reasoning over memorization~\citep{xu2024benchmarking,zhou2025lessleak}. Dynamic benchmarks have emerged as one response~\citep{kiela2021dynabench,shirali2022theory}, including continuously updated datasets such as LatestEval~\citep{li2024latesteval} and LiveBench~\citep{white2024livebench}. However, such approaches depend on sustained data curation and may be difficult to scale. Other paradigms, such as crowdsourced pairwise comparison in Chatbot Arena~\citep{chiang2024chatbot}, introduce variability driven by subjective or stylistic factors. Meanwhile, agent-based benchmarks~\citep{liu2023agentbench,zhou2023webarena,chan2024mle} primarily target domains with explicit verification signals, leaving the evaluation of general knowledge reasoning relatively underexplored. 

Puzzle tasks have become a widely used evaluation paradigm for LLMs. With clearly defined objectives and verifiable endpoints, they enable assessment not only of outcomes but also of trajectory-level signals of exploratory reasoning~\citep{giadikiaroglou2024puzzle}. Prior work has often focused on specific classes of puzzles~\citep{chollet2019measure,lin2021riddlesense,zhang2025puzzlebench}, including riddles, Sudoku, and related logic or combinatorial games. Recent benchmarks have advanced this approach. For instance, SPLAT~\citep{chen2024weak} frames tasks as clue-seeking within pre-scripted narratives, and MTP~\citep{badola2025multi} assesses models through carefully designed interactive environments. However, such environments typically define narrow action spaces that encourage models to depend on fixed strategies rather than engaging in deep knowledge reasoning. Moreover, the highly constrained goals of these scenarios restrict the evaluation method from being generalized to broader contexts. Our work extends the use of puzzle-solving by integrating multi-turn active probing with domain knowledge, aiming instead to bridge this gap and evaluate whether models can demonstrate a general understanding of any established concept.

\begin{figure}[t!]
\includegraphics[width=0.48\textwidth]{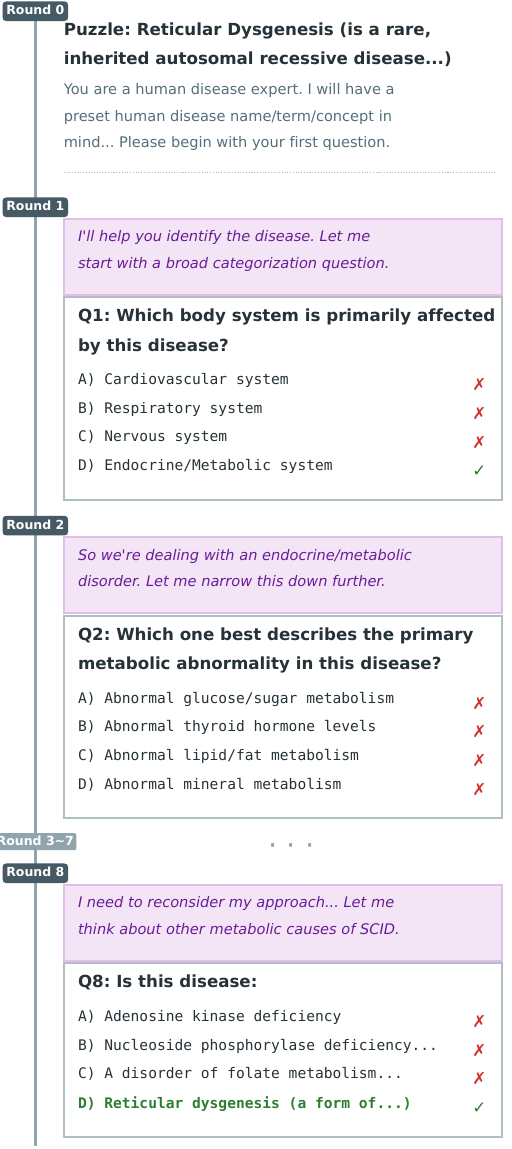}

\caption{Example episode from the Human Disease domain in APB, where the agent successfully identifies the target concept ``Reticular dysgenesis''. Across turns, the agent acquires diagnostic evidence through queries and feedback, progressively narrowing the hypothesis set until the target is determined. The resulting interaction trace provides process-level evidence of how the model accesses and applies specialized knowledge.}
\label{fig:sample_evaluation}
\end{figure}

In this work, we introduce the Active Probing Benchmark (APB), which expands the scope of puzzle-solving evaluation (Figure~\ref{fig:framework}). Specifically, APB assesses knowledge reasoning under interaction: the ability to identify a target concept through active multi-turn inquiry. This setup parallels practical decision making settings such as medical diagnosis, where a physician narrows possibilities through sequential questioning. Similarly, APB requires models to iteratively ask informative questions (Figure~\ref{fig:sample_evaluation}), use the returned signals to refine hypotheses, and efficiently reduce the candidate space until the target concept is identified. Since Wikipedia is a major source in the pre-training corpora of foundation models, APB constructs 960 puzzles centered on concepts of specialized difficulty spanning eight academic domains, including STEM, the humanities, and health sciences. By emphasizing information seeking interaction over fixed game mechanics, APB aims to evaluate how effectively LLMs can access and apply specialized knowledge in a controlled setting with explicit objectives.

Evaluation of a wide range of leading LLMs on APB reveals the benchmark's significant challenge where even the top-performing model \texttt{GPT-5} achieved a success rate of only 33.4\% (Table~\ref{tab:model_performance}). The results uncover a counter-intuitive phenomenon: most open-source reasoning-enhanced models significantly underperform their no-thinking counterparts, a trend exemplified by models including including \texttt{GPT-OSS-120B}, \texttt{DeepSeek R1}, and \texttt{Qwen3-235B-A22B-Thinking} (Table~\ref{tab:model_performance}). Detailed analysis of response patterns also highlights distinct failure modes in these models. For instance, Gemini's relative weakness stems from a tendency to frequently offer a conclusive guess which points to potential inherent flaws. Additionally, open source reasoning models tend to commit to specific candidate targets early and then probe in a confirmatory manner, leading to inefficient probing when the true target lies outside the initial hypothesis set. These analysis suggests that in interactive settings, the models struggle to leverage procedural reasoning to reliably access and apply specialized domain knowledge.

Beyond these empirical findings, the construction of the APB introduces a scalable method to rapidly generate diagnostic benchmarks in specific specialized domain, enabling a robust evaluation of knowledge reasoning. We further showcase its application by building an IC Back-End Puzzle Benchmark, featuring 480 puzzles from four EDA command tools in the IC back-end design (Appendix~\ref{app:sec:construction_method}). The core contribution can be summarized as follows:
\begin{itemize}[leftmargin=1.8em] 
\vspace{-3pt} 
\item We generalize the paradigm of puzzle-solving assessment by expanding the action and target spaces, shifting the focus from game mechanics to general knowledge reasoning across broader domains.
\item We introduce the Active Probing Benchmark (APB) and a scalable pipeline for generating evaluations of specialized difficulty, instantiated by 960 puzzles derived from expert-level concepts. 
\item We identify a critical proficiency gap in leading open-source reasoning models, observing that under the interactive evaluation paradigm of APB, these models struggle to leverage procedural reasoning to reliably access and apply specialized domain knowledge. \end{itemize}
\section{Related Work}
\subsection{Benchmark Evaluation}

The evaluation of large language models has been largely driven by a suite of comprehensive benchmarks designed to assess their knowledge and reasoning abilities across a multitude of domains~\citep{cobbe2021training,liu2024mathbench,rein2024gpqa,clark2018think,hendrycksmath2021}. Prominent examples such as MMLU~\citep{hendrycks2020measuring}, BIG-Bench~\citep{srivastava2023beyond} and HellaSwag~\citep{zellers2019hellaswag} have become standard tools for measuring model performance~\citep{contributors2023opencompass,duan2024vlmevalkit}. A common characteristic of these benchmarks is their reliance on static, single-turn formats, typically involving multiple-choice questions or direct answer generation~\citep{wang2024mmlu,suzgun2022challenging,yue2024mmmu,lu2022learn}.
However, static benchmarks are vulnerable to data leakage, since portions of their test data may inadvertently be included in the massive datasets used for model pre-training~\citep{balloccu2024leak,xu2024benchmarking,ni2025training,li2023open}.
One way to mitigate cheating is through dynamically updated benchmarks, such as LatestEval~\citep{li2024latesteval} and LiveBench~\citep{white2024livebench}, which continuously incorporate the latest real-world data. While this approach alleviates the issue to some extent, it still fails to enable fine-grained analysis and precise evaluation of the internal knowledge a model has acquired. To address this, we propose a dynamic, multi-turn framework designed to preclude rote memorization and compel the model to externalize its deductive reasoning.

\subsection{Puzzles and Games as Evaluation Paradigms}
Puzzles, broadly defined as structured problems that require reasoning to reach a solution, have long served as effective probes of intelligence~\citep{guo2024can,estermann2024puzzles,shah2024causal}. They constrain the search space while still demanding creativity, logical inference, and the integration of disparate knowledge. Accordingly, puzzles and games have been widely adopted as formal evaluation paradigms, from classic board games like Chess and Go~\citep{silver2017mastering} to text-based adventure games~\citep{wang2024can,gallotta2024large,zhang2025learning} and interactive formats such as 20 Questions~\citep{vinyals2019alphastar,liang2023leveraging,hu2018playing}. A common feature of these settings is sustained, multi-turn interaction within a structured environment, which makes them well-suited for assessing strategic planning, logical deduction, and efficient use of information~\citep{chen2024weak,li2025solving,chen2025finereason}. Motivated by this tradition, we frame evaluation as a multi-turn process that requires a model to externalize its reasoning trajectory rather than merely produce final answers. The Active Probing framework repurposes the puzzle format to evaluate how a model retrieves, integrates, and reasons over its internalized knowledge to identify a target concept it is expected to know, thereby prioritizing the knowledge structure and logical coherence of the reasoning trajectory over simple binary success metrics.

\section{Method}

\begin{figure*}[t!]
\centering
\includegraphics[width=0.82\textwidth]{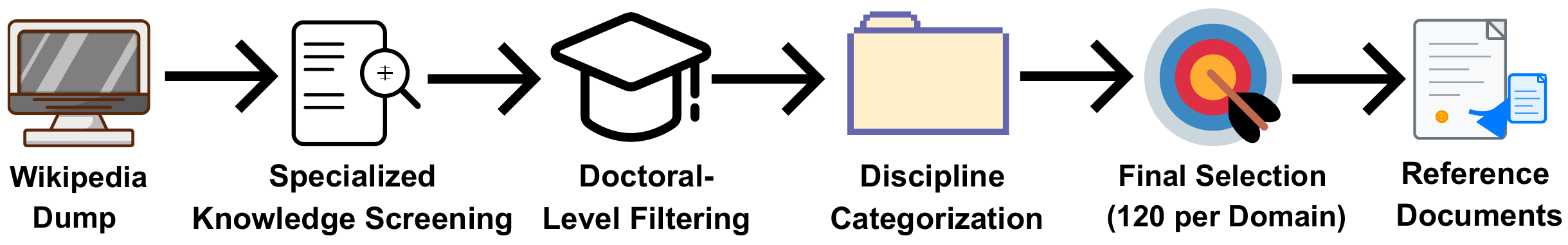}
\caption{The multi-stage, coarse-to-fine data filtering pipeline used to construct the Active Probing Benchmark from the English Wikipedia dump.}
\label{fig:data_pipeline}
\end{figure*}
\textsc{Active Probing Benchmark} is founded on a fundamental generalization of the puzzle definition, which enables the rapid construction of custom reasoning benchmarks. Section~\ref{sec:framework} details this generalized puzzle setting, along with our evaluation settings and metrics. Section~\ref{sec:benchmark} then describes the construction details in the APB.
\subsection{Multi-turn Active Probing Process}
\label{sec:framework}
Our evaluation method is a multi-turn interactive protocol. Its extensibility comes from a general definition of a puzzle as target concept identification through active evidence acquisition.

\paragraph{Generalized Puzzle Definition.}
Each instance specifies a hidden target concept $P_{\text{target}}$ (e.g., a specialized Wikipedia title or a technical command) and an associated reference document $I_{\text{ref}}$ that is accessible only to the oracle. The model must identify $P_{\text{target}}$ by asking questions and integrating turn-level feedback rather than by reading $I_{\text{ref}}$ directly.

\noindent\textbf{Problem Formulation.}
Let $V$ denote the concept space and initialize the candidate set as $S_0 = V$. The model interacts for at most $T=\texttt{MAX\_ROUNDS}$ turns. At each turn $t$, the model produces a multiple-choice query $Q_t=(q_t, O_t)$ and options $Q_t=(q_t, O_t)$ including question $q_t$ and options $O_t$, and receives feedback $A_t$. The interaction history is $H_t=\{(Q_1,A_1),\dots,(Q_t,A_t)\}$.

\noindent\textbf{Interaction Mechanism.}
Algorithm~\ref{alg:apb_episode} summarizes the interaction loop. At each turn $t$, the Querying Agent conditions on the accumulated history $H_{t-1}$ and outputs a multiple-choice query $Q_t=(q_t,O_t)$. If $P_{\text{target}} \in Q_t$, the episode terminates with \texttt{SUCCESS}; otherwise, the Knowledge Oracle consults the reference $I_{\text{ref}}$ and returns turn-level validation feedback $A_t=\mathcal{V}(Q_t,I_{\text{ref}})$ without exposing the reference text. The pair $(Q_t,A_t)$ is appended to form $H_t$ for the next turn, and the process continues until success or the budget $T$ is exhausted.

\begin{algorithm}[t]
\caption{APB multi-turn active probing process}
\label{alg:apb_episode}
\small
\begin{algorithmic}[1]
\REQUIRE Round budget $T$, puzzle target $P_{\text{target}}$, reference $I_{\text{ref}}$

\STATE $H_0 \leftarrow \emptyset$

\FOR{$t = 1$ to $T$}
    \STATE $Q_t=(q_t, O_t) \leftarrow \pi_{\theta}(H_{t-1})$ \COMMENT{Querying Agent: input $H_{t-1}$, output $Q_t$}
    
    \IF{$P_{\text{target}} \in Q_t$}
        \STATE \textbf{return} \texttt{SUCCESS}
    \ENDIF 
    
    \STATE $A_t \leftarrow \mathcal{V}(Q_t, I_{\text{ref}})$ \COMMENT{Knowledge Oracle: option validation feedback}
    \STATE $H_t \leftarrow H_{t-1} \cup \{(Q_t, A_t)\}$
\ENDFOR 

\STATE \textbf{return} \texttt{FAILURE}
\end{algorithmic}
\end{algorithm}

\subsection{The Active Probing Benchmark}
\label{sec:benchmark}
To assess the specialized knowledge that models internalize from core training corpora like Wikipedia, we constructed the \textbf{Active Probing Benchmark}, featuring 960 puzzles derived from Wikipedia across eight distinct domains.

\paragraph{Data Construction.}
The benchmark targets were curated via a rigorous, multi stage filtering pipeline (Figure~\ref{fig:data_pipeline}) that distills highly specialized concepts from a dump of 6.4M English Wikipedia titles. The pipeline acts as a coarse to fine funnel: it first removes broad classes of common entities and then progressively concentrates on concepts that require advanced domain expertise. The selection process consists of the following four key stages:
\begin{itemize}[nosep, leftmargin=*]
\item{\textbf{Stage 1 (Broad filter):}} We rapidly removed common entities, pop culture references, and general knowledge terms, yielding an initial pool of candidates.
\item{\textbf{Stage 2 (Expertise filter):}} We applied prompt based criteria that retain only titles judged to require doctorate level specialized knowledge claimed in prompts, discarding undergraduate or introductory concepts.
\item{\textbf{Stage 3 (Domain stratification):}} The remaining candidates were categorized into eight academic domains to ensure balanced coverage (e.g., biology, physics, humanities, and health sciences).
\item{\textbf{Stage 4 (Difficulty selection):}} Within each domain, we ran a dynamic tournament style selection: from batches of roughly 150 specialized candidates, we retained only the 5--10 most challenging items according to model based difficulty scoring. This step concentrates the final benchmark on the extreme tail of the difficulty distribution.
\end{itemize}
Across stages, the pipeline retains approximately 0.14\% of the original title set, producing a pool of roughly 9k advanced candidates and randomly selecting 120 targets per domain (960 puzzles total). The filtering code and prompts are released for reproducibility.\footnote{\url{https://anonymous.4open.science/r/wikipedia_filter-7870}} For each puzzle, the cleaned text from the corresponding Wikipedia article serves as the reference document $I_{\text{ref}}$.

\textbf{Difficulty validation.}
To provide quantitative evidence that the selected targets correspond to specialized, long tail knowledge, we analyze Wikipedia's link graph structure. For entities at different pipeline stages, we compute their in-link count (backlink count), which serves as a proxy for graph centrality (and correlates with commonness). As shown in Table~\ref{tab:topology_analysis}, entities rejected in Stage 1 exhibit saturated in-link counts (median $>500$, capped by the Wikipedia API), consistent with central and widely referenced concepts. In contrast, the final APB targets have much smaller in-link counts (median 16), indicating that they are substantially more peripheral in the knowledge graph and thus more specialized than the rejected groups.

\begin{table}[t]
\centering
\small 
\caption{Topological Analysis of Concept Difficulty. In-link counts drop drastically, confirming the selection of ``edge'' knowledge.}
\label{tab:topology_analysis}

\begin{tabularx}{\linewidth}{X c c c r}
\toprule
\textbf{Group} & \textbf{$N$} & \textbf{Mean} & \textbf{Med.} & \textbf{Type} \\
\midrule
G1: Rejected (Stage 1) & 1k & 379.2 & 500\textsuperscript{+} & Common \\
G2: Rejected (Stage 4) & 1k & 86.7 & 19.0 & Specialized \\
\textbf{G3: Final (APB)} & \textbf{1k} & \textbf{59.3} & \textbf{16.0} & \textbf{Deep/Edge} \\
\bottomrule
\end{tabularx}

\end{table}

\begin{figure}[t]
    \hspace*{-0.3cm}\includegraphics[width=0.53\textwidth]{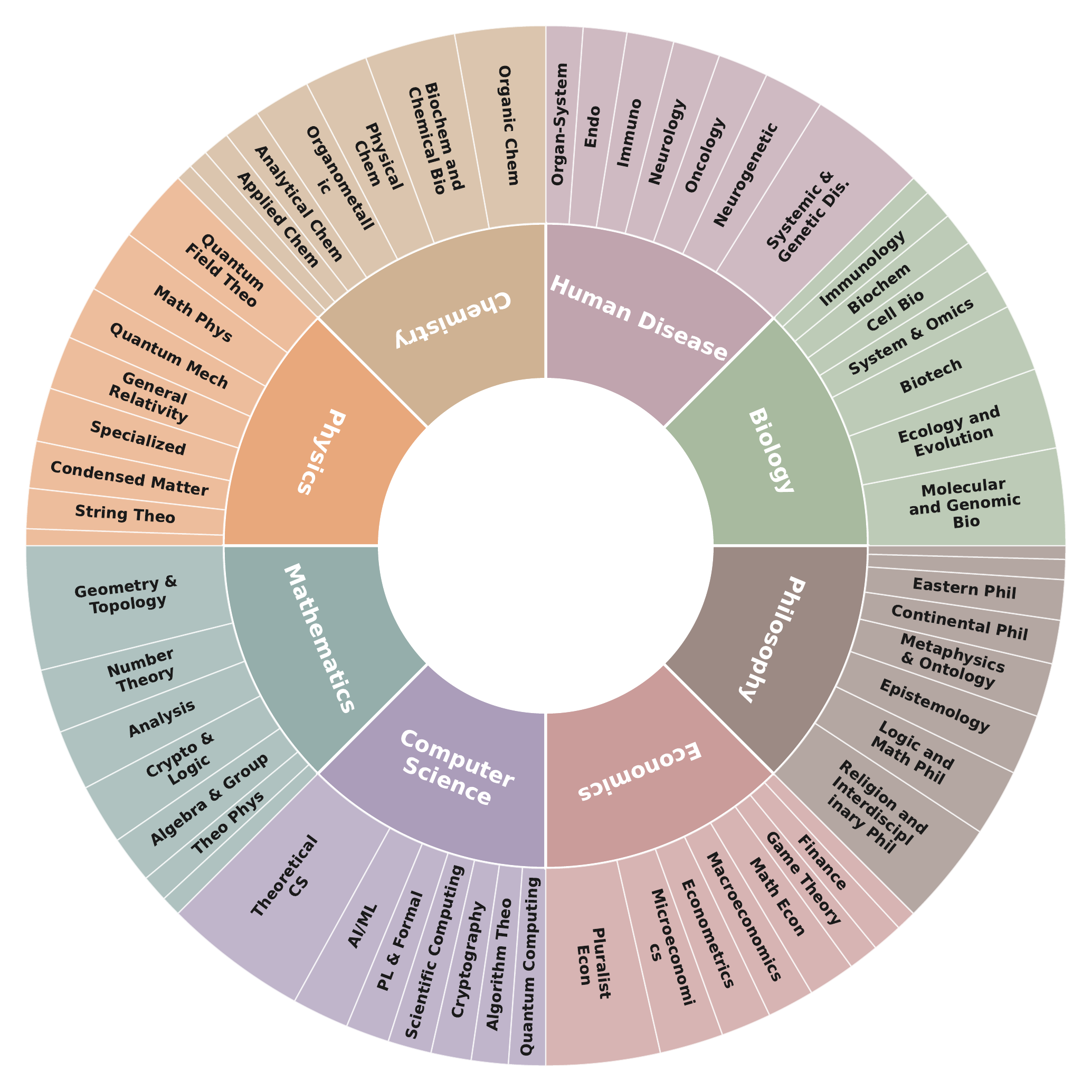}
    \caption{Hierarchical distribution of puzzle targets across domains and sub-disciplines. Eight distinct color families represent the primary domains, while individual segments denote specific sub-topics. The area of each segment is proportional to the number of questions in that category. Exact distributions are provided in Appendix~\ref{app:sec:detailed_metrics}.}
    \label{fig:benchmark_composition}
\end{figure}
\subsection{The Active Probing Benchmark}
\label{sec:benchmark_composition}
\paragraph{Benchmark Composition.}
The Active Probing Benchmark is composed of 960 puzzles, which are systematically categorized into eight distinct domains: Biology, Human Disease, Chemistry, Physics, Mathematics, Computer Science, Economics, and Philosophy. These domains were chosen to ensure comprehensive coverage of major academic fields, including the humanities and social sciences, the natural and engineering sciences, and health.

Each domain comprises 120 puzzles distributed across 7 to 8 primary sub-disciplines (Figure \ref{fig:benchmark_composition}) to ensure diverse topical coverage. For example, the Physics domain sources items from eight distinct subfields, with key contributions from Quantum Field Theory and Mathematical Physics. We also provide additional granularity within each subfield. For instance, the "Specialized Topics" sub-discipline covers Plasma Physics, Atomic, Molecular, and Optical (AMO) Physics, and Particle and Nuclear Physics. This multi-topic setup is a fundamental aspect of our benchmark across all domains, supporting a comprehensive evaluation of expert knowledge. Full details are described in the Appendix ~\ref{app:sec:detailed_metrics}.

\section{Experiments and Analyses}
\subsection{Setup}
This section outlines the key parameter configurations and the set of models evaluated in our experiments.
\begin{table*}
\centering
\caption{Success Rate (\%) for various models across eight domains. The best performance in each domain is highlighted in \textbf{bold}. The bottom row, 'Domain Avg.', summarizes the average performance for each domain.}
\label{tab:model_performance}
\small
\setlength{\tabcolsep}{5pt}
\renewcommand{\arraystretch}{1.2}
\begin{tabular}{lcccccccc|c}
\toprule
\textbf{Model} & \textbf{Bio} & \textbf{Dis} & \textbf{Chem} & \textbf{Phys} & \textbf{Math} & \textbf{CS} & \textbf{Econ} & \textbf{Phil} & \textbf{Avg.} \\
\midrule
GPT-5 & \textbf{34.2} & 37.5 & \textbf{33.3} & \textbf{34.2} & \textbf{34.2} & \textbf{25.0} & 39.2 & \textbf{30.0} & \textbf{33.4} \\
Claude Opus 4 & 27.5 & \textbf{40.8} & 29.2 & 26.7 & 18.3 & 19.2 & 31.7 & 27.5 & 27.6 \\
Claude Sonnet 4 & 29.2 & 35.0 & 23.3 & 23.3 & 17.5 & 18.3 & \textbf{42.5} & 21.7 & 26.4 \\
o3 & 28.3 & 33.3 & 21.7 & 29.2 & 20.0 & 10.0 & 35.0 & 25.0 & 25.3 \\
GPT-4o & 26.7 & 24.2 & 20.0 & 17.5 & 11.7 & 10.8 & 25.0 & 20.8 & 19.6 \\
Grok-4 & 14.2 & 30.8 & 16.7 & 13.3 & 10.0 & 13.3 & 20.8 & 21.7 & 17.6 \\
DeepSeek V3 & 21.7 & 22.5 & 13.3 & 15.0 & 11.7 & 11.7 & 18.3 & 16.7 & 16.4 \\
GPT-OSS-120B & 19.2 & 21.7 & 12.5 & 10.8 & 5.8 & 8.3 & 20.8 & 8.3 & 13.4 \\
Gemini 2.5 Pro No-Thinking & 9.2 & 28.3 & 12.5 & 6.7 & 5.8 & 7.5 & 16.7 & 17.5 & 13.0 \\
Qwen3-235B-A22B-Instruct & 5.8 & 27.5 & 5.0 & 11.7 & 5.8 & 10.0 & 15.0 & 20.0 & 12.6 \\
Gemini 2.5 Pro & 7.5 & 26.7 & 11.7 & 13.3 & 1.7 & 8.3 & 17.5 & 13.3 & 12.5 \\
DeepSeek R1 & 5.8 & 21.7 & 5.0 & 1.7 & 5.0 & 3.3 & 10.0 & 5.0 & 7.2 \\
Qwen3-235B-A22B-Thinking & 7.5 & 11.7 & 5.8 & 3.3 & 1.7 & 5.0 & 9.2 & 8.3 & 6.6 \\
\midrule
\textbf{Domain Avg.} & 18.2 & \textbf{27.8} & 16.2 & 15.9 & \textbf{11.5} & 11.6 & 23.2 & 18.1 & \textbf{17.8} \\
\bottomrule
\end{tabular}
\end{table*}

\noindent\noindent\textbf{Maximum Rounds.}
We set \texttt{MAX\_ROUNDS} to 15. This decision is justified by empirical data (Table \ref{tab:detailed_model_metrics}), which shows that a successful session concludes in 9.78 turns on average (9.15 for the top-performing \texttt{GPT-5}). A limit of 15 thus provides sufficient opportunity for models to converge without being cost-prohibitive. Simultaneously, this cap manages the escalating token costs associated with including the full conversation history in each turn. While increasing the limit to 20 would have nearly doubled the evaluation cost.

\noindent\textbf{Temperature.} We standardized the temperature at 0.2 to ensure reproducibility. Our ablation study (Appendix~\ref{app:sec:ablation_params}) confirms the robustness of this choice, demonstrating that alternative settings (e.g., higher temperatures for reasoning models) yield negligible gains and do not alter the performance hierarchy. \texttt{GPT-5} used its default ``medium'' effort, verified as optimal in our analysis."

\noindent\textbf{Knowledge Oracle.} We designated \texttt{DeepSeek V3} as the Knowledge Oracle responsible for providing correctness feedback. This model was chosen for its strong instruction-following capabilities and consistent outputs at a low temperature. To ensure our main findings are not sensitive to this specific choice, we conducted a detailed ablation study (Appendix~\ref{app:sec:ablation_study}). This includes both a subset analysis and a comprehensive cross-validation using \texttt{Claude Sonnet 4} on the full 960-puzzle benchmark, confirming that the relative performance hierarchy of models remains stable ($\rho=0.96$) despite absolute scoring shifts.

\noindent\textbf{Evaluated Models.} Our evaluation on the APB covers a comprehensive suite of 13 leading closed-source and open-source LLMs to assess their capabilities. We list all the evaluated models below: 
\begin{itemize}
    \item \textbf{Closed-Source models}: \texttt{GPT-5}, \texttt{o3} and \texttt{GPT-4o}~\citep{hurst2024gpt}, \texttt{Claude Opus 4} and \texttt{Claude Sonnet 4}, \texttt{Gemini 2.5 Pro} and \texttt{Gemini 2.5 Pro No-Thinking}~\citep{comanici2025gemini}, \texttt{Grok-4}.
    \item \textbf{Open-Source models}: \texttt{Deepseek V3}~\citep{liu2024deepseek} and \texttt{Deepseek R1}~\citep{guo2025deepseek}, \texttt{GPT-OSS-120B}~\citep{agarwal2025gpt}, \texttt{Qwen3-235B-A22B-Instruct} and \texttt{Qwen-235B-A22B-Thinking}~\citep{yang2025qwen3}.
\end{itemize}

\subsection{Main Results}
The main evaluation results are presented in Table~\ref{tab:model_performance}, covering detailed success rate assessments across 8 domains: Biology (Bio), Human Disease (Dis), Chemistry (Chem), Physics (Phys), Mathematics (Math), Computer Science (CS), Economics (Econ), and Philosophy (Phil).

\noindent\textbf{Analysis of Main Results.} The evaluation results demonstrate pronounced disparities across 8 domains (Table~\ref{tab:model_performance}). In the strongest-performing domains, Human Disease (27.8\%) and Economics (23.2\%) achieved the highest average success rates. The high accuracy in Human Disease can be attributed to its comprehensive and well-structured classification system, while the strong performance in Economics may stem from the relatively lower depth and complexity of its specialized knowledge. In the most challenging domains, Mathematics (11.5\%) and Computer Science (11.6\%) recorded the lowest success rates, where even \texttt{GPT-5}, the strongest model, attained only 34.2\% and 25.0\% respectively. Intermediate performance was observed in Biology, Chemistry, Physics, and Philosophy.

From a model-centric perspective, the results reveal a clear performance hierarchy. While \texttt{GPT-5} leads the pack, its predecessor \texttt{GPT-4o} (19.6\%) demonstrates remarkable longevity, effectively separating the top-performing models and the rest. When compared to other closed-source offerings, the relatively modest results from the \texttt{Gemini} series point to specific weaknesses for this reasoning task. Furthermore, the struggles of \texttt{DeepSeek R1} (7.2\%) and \texttt{Qwen3-235B-A22B-Thinking} (6.6\%) underscore the current limitations and vulnerabilities within the open-source reasoning model landscape.

\begin{table}[t]
    \centering
    \includegraphics[width=1\linewidth]{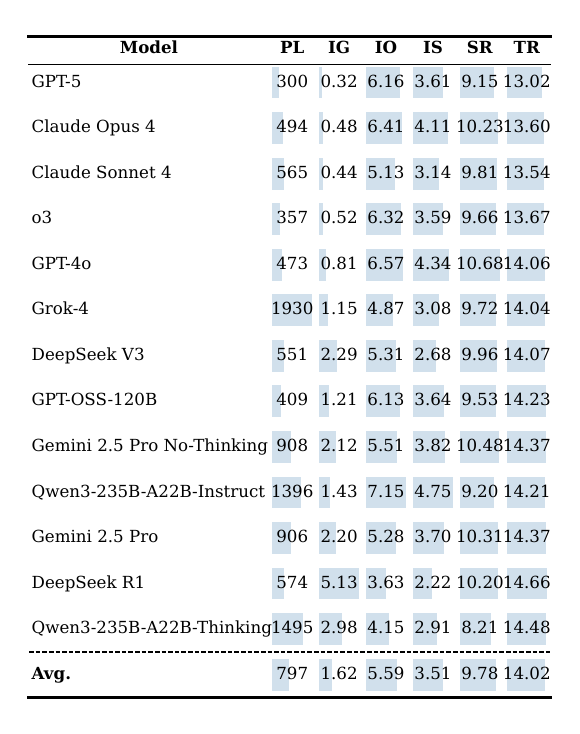}
    \caption{Detailed behavioral analysis metrics. \textbf{PL}: Probing Length, \textbf{IG}: Incorrect Guess rounds, \textbf{IO}: Incorrect Option rounds, \textbf{IS}: Incorrect Streak rounds, \textbf{SR}: Success Rounds, \textbf{TR}: Total Rounds. Blue bars visualize the normalized magnitude of each metric within its column to highlight behavioral patterns.}
    
    \label{tab:detailed_model_metrics}
\end{table}

\subsection{Detailed Analyses}
This section provides a detailed analysis of model performance based on Table~\ref{tab:detailed_model_metrics}, focusing on the behavioral patterns observed during the knowledge reasoning process.

\noindent\textbf{Probing Length.}
The Probing Length (PL) measures the average character length of the agent's output per turn. This output typically includes not only the multiple-choice question but also the model's explicit reasoning based on prior rounds. Our analysis reveals that increased verbosity does not correlate with better performance on this task. Indeed, efficiency in knowledge probing rounds appears to be a hallmark of the top models like \texttt{GPT-5} (PL = 300). This trend generally holds for the other top-performing models with relatively concise probing lengths. Furthermore, we observe that PL is often strongly correlated with a model's developer family, suggesting that different series of models possess inherent and consistent styles for expressing their reasoning process in this interactive setting. For instance, both \texttt{Gemini} models exhibit a similarly high verbosity (PL $\approx$ 907), as do the \texttt{DeepSeek} models (PL $\approx$ 560) and the \texttt{OpenAI} models ($299 < \text{PL} < 500$). This suggests that different series of models possess inherent and consistent styles for expressing their probing process in this interactive setting.

\noindent\textbf{Incorrect Guess Rounds.} The Incorrect Guess Rounds (IG) metric quantifies the average number of times an agent makes a conclusive but incorrect guess of the target concept per session. We interpret a high IG as an indicator of model overconfidence, where the agent prematurely commits to a wrong answer despite having insufficient information. Our results show a strong negative correlation between IG and Success Rate, suggesting that early incorrect commitments are associated with persistent mislocalization in the knowledge space, which limits the usefulness of subsequent probing. For instance, the top-performing model \texttt{GPT-5} has the lowest IG (0.32), while the struggling \texttt{DeepSeek R1} has the highest (5.13). From a strategic standpoint, such actions represent a low-utility approach by invalidating only a single hypothesis. In a context like Human Disease, a propensity for making frequent, incorrect assertions is a harmful failure mode that could lead to detrimental outcomes, underscoring the importance of penalizing such overconfident behavior in model evaluation.

\noindent\textbf{Incorrect Probing and Error Correction.} We analyze two metrics related to probing errors: Incorrect Option rounds (IO) and Incorrect Streak rounds (IS). IO measures the average number of rounds where all four proposed options are incorrect, suggesting inaccurate concept grounding during knowledge localization. IS measures the longest consecutive sequence of such rounds, reflecting the model's ability to recover from a flawed reasoning path. However, these metrics are not simple indicators of poor performance. An 'all-incorrect' round can be highly informative for a capable agent, as it allows for significant pruning of the search space and a major strategy correction. This can paradoxically help a strong model quickly locate an obscure concept. Thus, these metrics offer insights into a model's strategic behavior. The case of \texttt{DeepSeek R1} is particularly illustrative: despite having the lowest IO (3.63) and IS (2.22), it achieves one of the lowest success rates. This suggests that its probing strategy is ineffective at leveraging information and asking questions to find the $P_{\text{target}}$, a strategic failure that helps explain its high rate of premature guesses, as reflected in its top-ranking IG.

\noindent\textbf{Overall Efficiency.} We evaluate efficiency using two complementary metrics: Average Success Rounds (SR) and Average Total Rounds (TR). SR measures the average number of rounds in successful sessions only, reflecting speed on solvable problems. TR provides a more holistic score by including failed sessions, which are counted as \texttt{MAX\_ROUNDS} (15), thus blending success rate with efficiency. Relying solely on SR can be misleading for models with low success rates, as it only reflects performance on easier tasks they could solve. TR, therefore, offers a more comprehensive view. For example, \texttt{GPT-5}'s top performance gives it the best TR (13.02) and leading SR (9.15), confirming its position as the most effective model overall. This combined metric also reveals nuances in reasoning efficiency. For instance, despite \texttt{Claude Opus 4} having a higher Success Rate than \texttt{Claude Sonnet 4}, \texttt{Sonnet}'s lower TR (13.54 vs. 13.60) and SR (9.81 vs 10.23) suggest it has a relative advantage in the efficiency of its reasoning process.

\subsection{Error Analysis}
\begin{figure}[t]
\centering
\includegraphics[width=0.8\linewidth]{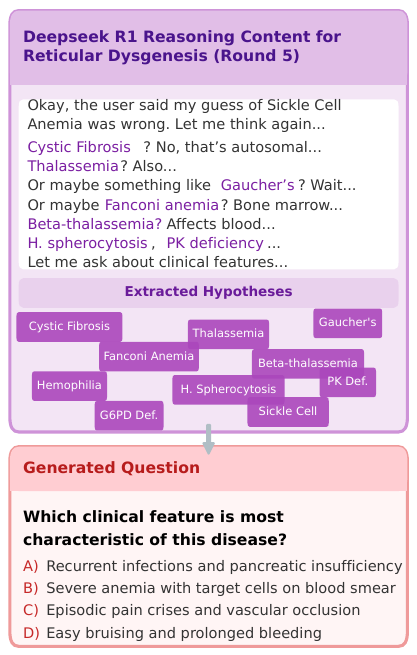}
\caption{\texttt{DeepSeek R1} reasoning trace for the target 'Reticular dysgenesis' (Round 5). \textbf{Top:} The internal reasoning content shows the model brainstorming specific disease candidates following an incorrect guess. \textbf{Middle:} The extracted hypotheses visualize the diverse set of conditions considered during this step. \textbf{Bottom:} The generated follow-up question focuses on differentiating clinical features.}
\label{fig:deepseek_reasoning_sample}
\end{figure}

\noindent\textbf{Open-Source Reasoning Models.}
To understand the relatively weak performance of open source reasoning models, we analyzed the publicly available chain-of-thought traces for \texttt{GPT-OSS-120B}, \texttt{DeepSeek R1}, and \texttt{Qwen3-235B-A22B-Thinking}. We observe a recurring pattern that is best characterized as a failure of knowledge localization during active probing. Rather than using early turns to acquire broadly discriminative evidence, these models often anchor on several concrete candidate targets for $P_{\text{target}}$ at the beginning of an episode (Figure~\ref{fig:deepseek_reasoning_sample}). Subsequent multiple-choice queries are then constructed to seek confirmation for these candidates, instead of eliciting information that would re-partition the broader candidate space.

This premature anchoring has two downstream effects. First, when the true $P_{\text{target}}$ is not among the initial hypotheses, the model's questions tend to target attributes that are irrelevant to the correct region of the domain knowledge space, yielding feedback that is weakly informative for correction. As a result, the interaction trace exhibits repetitive cycles of querying around the wrong neighborhood, with many low-utility terms and little reduction of the remaining uncertainty. Second, the same anchoring increases the likelihood of conclusive but unsupported commitments, producing more frequent incorrect final guesses. This is consistent with the comparatively high IG values observed for these models and with their lower success rates overall.

These traces suggest that the performance gap is driven less by a lack of procedural reasoning steps and more by difficulty in using multi-turn feedback to maintain and update a well-grounded hypothesis set. In other words, the models often fail to treat early turns as evidence acquisition for concept identification, and instead collapse prematurely onto specific candidates, which limits their ability to recover when the initial localization is wrong.

\noindent\textbf{Gemini Series Models.}
The \texttt{Gemini 2.5 Pro} models exhibit a distinct behavioral pattern that helps explain their relatively weak performance. A primary characteristic is their high Probing Length (PL $\approx$ 907), far exceeding other top closed-source models. We observe that this verbosity stems from a tendency to repeatedly summarize and elaborate on prior turns before issuing the next query. From a knowledge probing perspective, this behavior shifts effort from acquiring new discriminative evidence to reprocessing self-generated context, which can dilute the influence of objective turn-level feedback. As a result, the model may persist in a mislocalized hypothesis set and continue probing within an irrelevant region of the knowledge space, culminating in more frequent unsupported final commitments, reflected by their notably high IG (2.1--2.2) relative to other leading closed-source models.

\section{Conclusion}

We introduced APB, an interactive benchmark that extends puzzle evaluation from narrow game-like settings to knowledge reasoning over specialized concepts. By casting concept identification as multi-turn information seeking with verifiable endpoints, APB provides process-level evidence of how models acquire and use domain knowledge. Experiments show that APB is challenging even for strong systems (\texttt{GPT-5} reaches 33.4\% success), and that several open source reasoning models lag behind their no-thinking counterparts, consistent with difficulties in grounding multi-turn inference in specialized knowledge. Beyond these results, APB offers a practical construction pipeline for producing challenging diagnostic evaluations in additional specialized domains, helping broaden the empirical study of knowledge reasoning under interaction.
\section*{Impact Statement}
The primary objective of this work is to evaluate and better understand the limitations of interactive knowledge reasoning in large language models, contributing to more trustworthy and informative model assessment. APB is intended for research use and reports objective success criteria together with qualitative failure analyses to reduce over-interpretation of single metrics. This study was conducted independently and adheres to standard ethical research principles, including considerations of bias, fairness, privacy, security, and legal compliance.
\bibliography{example_paper}
\bibliographystyle{icml2026}
\appendix
\label{sec:appendix}
\clearpage
\appendix
\section*{Appendix}

\section{Rapid Custom Benchmark Construction}
\label{app:sec:construction_method}
Our framework provides a scalable and generalizable method for rapidly constructing diagnostic benchmarks in any specialized domain. The process is straightforward and consists of two main stages:
\begin{itemize}[leftmargin=1.8em]
\vspace{-3pt}
    \item \textbf{Target Identification and Curation}: First, a domain of interest and a corresponding source corpus of knowledge are identified (e.g., technical manuals, API documentation, academic articles). From this corpus, a list of key concepts, terms, commands, or entities is systematically extracted to serve as the puzzle targets ($P_{\text{target}}$)

    \item \textbf{Reference Document Generation}: For each puzzle target, a "golden" reference document ($I_\text{ref}$) is established (e.g., the specific documentation page for a command). This document serves as the ground truth for the environment's feedback during the multi-turn evaluation.
\end{itemize}
\subsection{Case Study: The IC Back-End Puzzle Benchmark}

To demonstrate the efficacy of this method, we applied it to the niche, industrial domain of IC back-end design. We used the official documentation for four widely-used EDA command sets—Synopsys ICC, ICC2, and Cadence Innovus (both common and legacy)—as our source corpus.

\begin{table}[h]
\centering
\small
\setlength{\tabcolsep}{3.5pt} 
\renewcommand{\arraystretch}{1.15}

\caption{Success rates (\%) on the IC Back-End Puzzle Benchmark. Tools are grouped by vendor: Synopsys (ICC, ICC2) and Cadence Innovus (Common, Legacy). ``NT'' denotes ``No-Thinking''. \textbf{Qwen3 models refer to the Qwen3-235B-A22B variants.} Best results in each category are \textbf{bolded}.}
\label{tab:ic_backend_performance}

\begin{tabular*}{\linewidth}{l @{\extracolsep{\fill}} cccc|c}
\toprule
 & \multicolumn{2}{c}{\textbf{Synopsys}} & \multicolumn{2}{c}{\textbf{Cadence Innovus}} & \\
\cmidrule(lr){2-3} \cmidrule(lr){4-5} 
\textbf{Model} & \textbf{ICC} & \textbf{ICC2} & \textbf{Common} & \textbf{Legacy} & \textbf{Avg.} \\
\midrule
Claude Opus 4 & \textbf{27.5} & 14.2 & 16.7 & 22.5 & \textbf{20.2} \\
GPT-5 & 25.8 & \textbf{15.0} & 14.2 & 20.0 & 18.8 \\
grok-4 & 15.0 & 10.8 & \textbf{18.3} & \textbf{24.2} & 17.1 \\
o3 & 20.0 & \textbf{15.0} & 12.5 & 12.5 & 15.0 \\
Claude Sonnet 4 & 14.2 & 10.8 & 9.2 & 18.3 & 13.1 \\
Gemini 2.5 Pro & 10.8 & 8.3 & 5.0 & 16.7 & 10.2 \\
Gemini 2.5 Pro NT & 10.8 & 8.3 & 6.7 & 12.5 & 9.6 \\
GPT-4o & 6.7 & 7.5 & 7.5 & 7.5 & 7.3 \\
DeepSeek V3 & 10.8 & 4.2 & 6.7 & 5.8 & 6.9 \\
GPT-OSS-120B & 8.3 & 7.5 & 4.2 & 4.2 & 6.0 \\
Qwen3 Instruct & 4.2 & 5.8 & 3.3 & 4.2 & 4.4 \\
Qwen3 Thinking & 6.7 & 2.5 & 3.3 & 5.0 & 4.4 \\
DeepSeek R1 & 4.2 & 0.8 & 2.5 & 2.5 & 2.5 \\
\midrule
\textbf{Domain Avg.} & 12.7 & 8.5 & 8.5 & 12.0 & \textbf{10.4} \\
\bottomrule
\end{tabular*}
\end{table}

Following our process, we curated 480 puzzles by sampling 120 command names from each of the four sets. This benchmark exemplifies how our framework can be used to quickly create robust, expert-level evaluations that test practical knowledge in any specialized domain.

We evaluated 13 leading LLMs on this benchmark, with the results presented in Table \ref{tab:ic_backend_performance}. The findings underscore the benchmark's significant challenge, with an average success rate of only 10.4\% across all models. A clear performance hierarchy emerged, with Claude Opus 4 (20.2\%), GPT-5 (18.8\%), and Grok-4 (17.1\%) forming a distinct top tier. Notably, reasoning-enhanced models like DeepSeek R1 (2.5\%) and Qwen3 235B Thinking (4.4\%) performed very poorly, reinforcing our main paper's finding of a gap between procedural reasoning and applied domain knowledge. The domain-specific results also reveal non-uniform knowledge; for instance, Grok-4 excels on Innovus commands but is weaker on ICC, suggesting that model knowledge of these specialized tools is often fragmented. This case study validates our method's ability to create challenging benchmarks that effectively differentiate model capabilities in highly specialized, real-world domains.

\subsection{Broader Applications and Future Directions}
The methodology detailed in this section is designed for broad applicability beyond the technical domains demonstrated. We encourage the research community to leverage our scalable process to construct new benchmarks across a wide array of fields. For example, puzzle targets could be drawn from the humanities (e.g., famous figures, artworks, film titles) or the natural sciences (e.g., celestial bodies, materials, biological species).

The creation of such diverse benchmarks would facilitate a more holistic evaluation of an LLM's all-around knowledge reasoning abilities. Moreover, the resulting multi-turn interaction data would provide a rich resource for deeper, cross-domain pattern recognition, enabling a more comprehensive exploration of model reasoning strategies and behaviors.
\begin{figure*}[h]
\centering
\hspace*{-0.8cm}\includegraphics[width=0.8\linewidth]{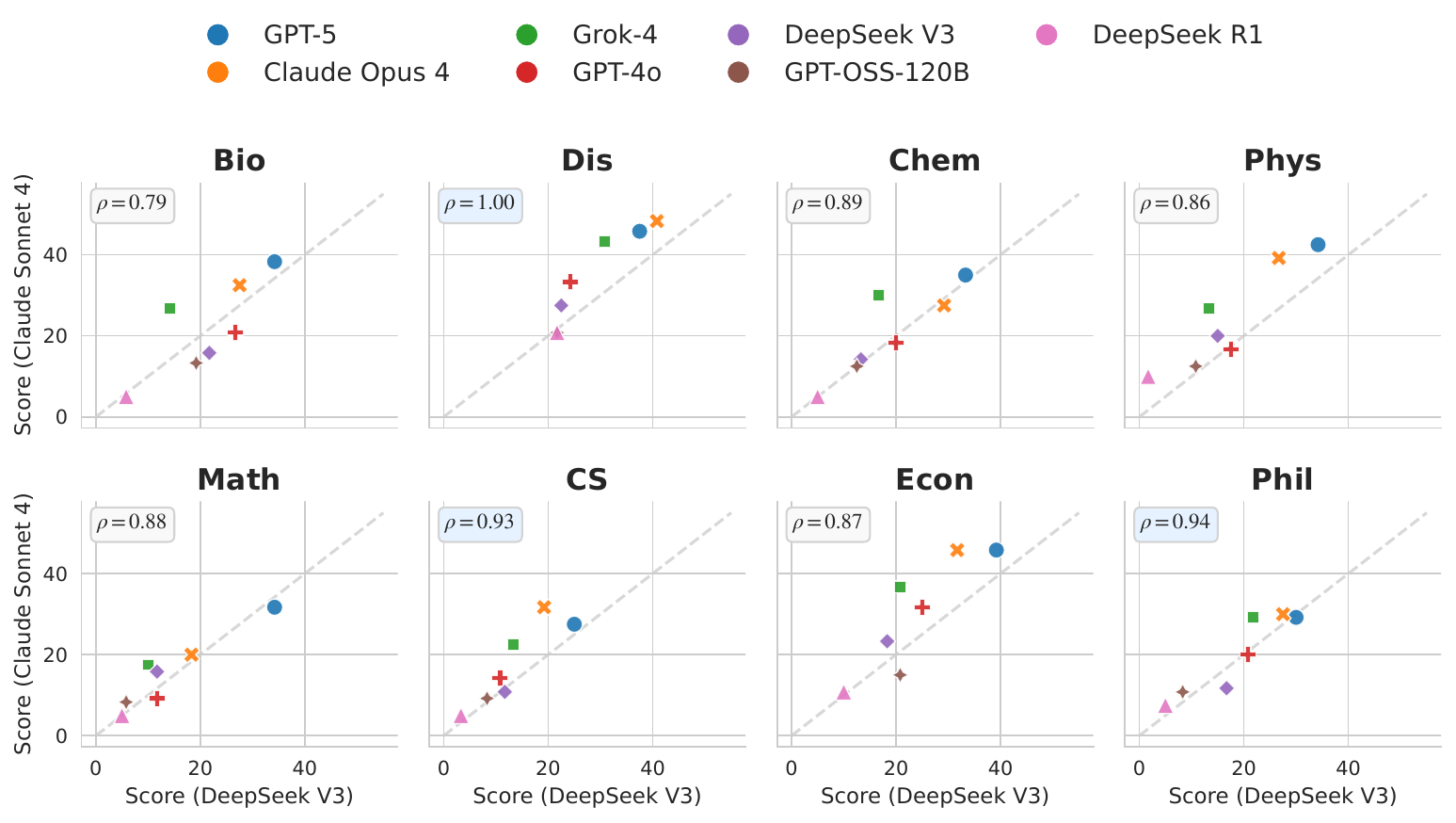} 
\caption{Ranking consistency analysis (DeepSeek V3 vs. Claude Sonnet 4) across eight domains. Each point represents a model's performance in a specific domain. The distinct clustering of model colors and the high Spearman's rank correlation ($\rho$) within each domain confirm that relative model rankings remain robust despite the absolute score shift caused by the stricter judge.}
\label{fig:ranking_consistency}
\end{figure*}
\section{Ablation Study}
\label{app:sec:ablation}

To validate the robustness of our evaluation framework, we conducted a comprehensive ablation study identifying potential sources of variance in our experimental pipeline. Our objective is to confirm that the observed model performance tiers are intrinsic to the agents' capabilities rather than artifacts of the evaluation configuration. Specifically, we address two critical dimensions: (1) the reliability of the \textbf{Knowledge Oracle} (Judge Consistency), and (2) the sensitivity of performance to \textbf{hyperparameters} and inference-time compute (Temperature and Reasoning Effort).

\subsection{Cross-Validation with Claude Sonnet 4}
\label{app:sec:ablation_study}
We employed Claude Sonnet 4 to cross-validate the main results against DeepSeek V3. This evaluation covered the full core benchmark (960 puzzles) across seven representative models, covering the spectrum from top-tier proprietary models to open-source and reasoning-specific models.

\paragraph{Rank Stability Across Different Judges}
Figure~\ref{fig:ranking_consistency} visualizes the domain-level performance correlation between the two judges. While absolute success rates tend to be higher under Claude Sonnet 4 (as indicated by the data points generally lying above the $y=x$ diagonal), the relative performance ordering remains remarkably stable:

\begin{itemize}[leftmargin=1.5em]
\item \textbf{Visual Consistency:} As shown in the facet plots, models form distinct clusters. Top-performing models (e.g., GPT-5, Claude Opus 4) consistently appear in the upper-right quadrant, while lower-tier models remain in the lower-left across all domains.
\item \textbf{High Correlation:} The quantitative evidence supports this visual trend. We observe a Spearman rank correlation of $\rho=0.96$ for the overall model rankings, indicating near-identical leaderboard structures. This consistency remains robust at the granular level, with an average correlation of $\rho=0.89$ across individual domains (exemplified by $\rho=1.00$ in Disease and $\rho=0.94$ in Philosophy).
\item \textbf{Robust Failure Modes:} ``Reasoning-enhanced'' models (DeepSeek R1, GPT-OSS-120B) remain at the bottom regardless of the judge. Notably, even with the more capable Claude Sonnet 4 as the judge, DeepSeek R1 achieves only 8.6\% accuracy compared to 7.2\% under V3. This confirms that the failure of these models stems from their internal strategies (i.e., the concrete-to-abstract fallacy), not judge-induced noise.
\end{itemize}

\paragraph{Explanation for Spikes and Judge Differences}
Our case analysis identified two primary factors driving the observed differences between judges: context robustness and semantic validation style.

\begin{itemize}[leftmargin=1.5em]
\item \textbf{Grok-4's Outlier Performance (Context Robustness):} Grok-4 exhibits an anomalously high probing length (avg. 1930 characters/turn). DeepSeek V3 struggles to maintain precise instruction-following when parsing such verbose contexts. In contrast, Claude Sonnet 4 effectively filters through Grok-4's lengthy outputs to extract the core logic, explaining why Grok-4 shows the most dramatic improvement (+11.5\%) under the stronger judge.
\item \textbf{Judging Characteristics (Association vs. Precision):} DeepSeek V3 tends toward \textit{associative validation}—occasionally accepting highly related ``attributes'' as the target ``entity'' itself. In contrast, Claude Sonnet 4 enforces stricter semantic precision.
\end{itemize}

In conclusion, replacing DeepSeek V3 with Claude Sonnet 4 acts primarily as a \textbf{linear scaling factor}. It uniformly ``lifts'' the scores of capable models but does not fundamentally alter the leaderboard. The high rank correlation confirms that DeepSeek V3 remains a valid, cost-effective proxy for differentiating model tiers.

\subsection{Robustness to Sampling Parameters and Reasoning Effort}
\label{app:sec:ablation_params}

To ensure that our main findings are not artifacts of suboptimal hyperparameters, we conducted a targeted ablation study on a representative subset of the benchmark ($N=120$). We isolate the effects of sampling temperature on open-source reasoning models and the impact of test-time compute scaling on proprietary models.

\paragraph{Temperature Sensitivity in Open-Source Models}
A potential hypothesis for the underperformance of open-source reasoning models (e.g., DeepSeek R1, Qwen3) is that the default low-temperature settings ($T=0.2$) might constrain their exploratory capabilities. To test this, we evaluated DeepSeek R1 at its recommended high temperature ($T=1.0$) and Qwen3 Thinking at a moderate setting ($T=0.6$).

As detailed in Table~\ref{tab:ablation_params}, the results indicate that temperature is not the bottleneck. 
\begin{itemize}[leftmargin=1.5em]
    \item \textbf{DeepSeek R1:} While increasing the temperature to $T=1.0$ yielded a marginal improvement (6.7\% $\to$ 8.3\%), the model remains firmly in the bottom tier.
    \item \textbf{Qwen3:} Increasing the temperature resulted in a slight performance fluctuation (11.7\% $\to$ 10.8\%), offering no statistical benefit.
\end{itemize}
These findings reinforce our conclusion that the failure modes of these models—specifically the concrete-to-abstract fallacy—are structural rather than parameter-dependent.

\begin{table}[h]
\centering
\caption{Ablation results on sampling temperature and reasoning effort ($N=120$). The ``Tier'' column reflects the relative ranking within the full benchmark context.}
\label{tab:ablation_params}
\resizebox{0.9\linewidth}{!}{
\begin{tabular}{llcc}
\toprule
\textbf{Model} & \textbf{Configuration} & \textbf{Success Rate} & \textbf{Tier} \\ 
\midrule
\multicolumn{4}{l}{\textit{Effect of Temperature (Open-Source Reasoning)}} \\
\midrule
\multirow{2}{*}{DeepSeek R1} & Temp = 0.2 (Default) & 6.7\% & Bottom \\
 & Temp = 1.0 (Recommended) & 8.3\% & Bottom \\ 
\cmidrule{2-4}
\multirow{2}{*}{Qwen3 Thinking} & Temp = 0.2 (Default) & 11.7\% & Low \\
 & Temp = 0.6 & 10.8\% & Low \\
\midrule
\multicolumn{4}{l}{\textit{Effect of Reasoning Effort (GPT-5)}} \\
\midrule
\multirow{3}{*}{GPT-5} & Low Effort & 33.3\% & Top \\
 & Medium Effort (Default) & \textbf{38.3\%} & \textbf{Top} \\
 & High Effort & 34.6\% & Top \\
\bottomrule
\end{tabular}
}
\end{table}

\paragraph{Scaling Laws of Test-Time Compute}
We further investigated the relationship between reasoning effort and performance using GPT-5. By varying the inference budget across Low, Medium (Default), and High settings, we observed a distinct \textbf{inverted-U trajectory}: performance peaks at Medium effort (38.3\%) and declines at High effort (34.6\%).

This observation leads to two critical insights:
\begin{enumerate}[leftmargin=1.5em]
    \item \textbf{Diminishing Returns:} In the constrained paradigm of strategic questioning, excessive test-time compute does not linearly translate to better outcomes. "Over-thinking" may lead to hallucinated constraints or overly convoluted search paths.
    \item \textbf{Strategy Quality over Quantity:} Crucially, even at the ``Low'' effort setting, GPT-5 (33.3\%) significantly outperforms the best open-source reasoning models. This suggests that the \textit{quality} of the underlying reasoning strategy—how the model prunes the search space—is a more determinant factor than the raw \textit{quantity} of inference steps or token generation.
\end{enumerate}

\section{Token Usage}
\label{app:sec:token_usage}
This section details the token consumption patterns, averaged per puzzle, for both the Querying Agent (QA) and the Knowledge Oracle (KO). It is important to note that a single puzzle evaluation consists of multiple rounds of interaction.

The token cost of the Knowledge Oracle (KO) is characterized by highly efficient outputs but demanding inputs. The KO's completion tokens are minimal and stable across all tests (averaging 460 tokens), as its responses are restricted to a fixed format providing only correctness judgments. This ensures a fair and consistent evaluation. However, its prompt token requirements are substantial, averaging 18,347 tokens per puzzle, because the KO must process the full reference document ($I_\text{ref}$) in each round to provide accurate feedback.

Similarly, the QA prompt costs are significant, averaging 18,826 tokens per puzzle, as the full interaction history is included in the context for each new turn. The vast majority of performance variance and economic cost, however, comes from the SA’s completion tokens, which reflect model verbosity (Table \ref{tab:token_summary}). Notably, some of the poorest-performing open-source reasoning models were paradoxically the most computationally expensive. GPT-OSS-120B was the most costly, consuming over 72,000 total QA tokens on average, followed by Qwen3 235B Thinking (approx. 57,600 tokens) and DeepSeek R1 (approx. 43,300 tokens). Collectively, these three open-source models averaged 57,784 QA tokens per puzzle—almost exactly double the 28,705-token average of the remaining models. This contrasts sharply with top-performing models like GPT-5, which achieved the highest success rate while using less than half the tokens of GPT-OSS-120B. This suggests that excessive verbosity is not only an unsuccessful strategy but also an economically inefficient one for this task.
\vspace{0.5cm}
\begin{table}[H]
\centering
\caption{A summary of token usage for representative models. "Comp." stands for Completion, "Tok." for Tokens. KO output is low due to fixed formatting.}
\label{tab:token_summary}
\setlength{\tabcolsep}{3pt}
\resizebox{\linewidth}{!}{
\begin{tabular}{l|ccc|c}
\toprule
\textbf{Model} & \textbf{QA} & \textbf{QA} & \textbf{QA Total} & \textbf{KO Comp.} \\
 & \textbf{Prompt} & \textbf{Comp.} & \textbf{Tok.} & \textbf{Tok.} \\
\midrule
\multicolumn{5}{l}{\textit{(Best Performance)}} \\
GPT-5 & 9,821 & 24,301 & 34,122 & 430 \\
\midrule
\multicolumn{5}{l}{\textit{(Most Verbosity)}} \\
GPT-OSS-120B & 14,034 & 58,471 & 72,505 & 473 \\
Qwen3 235B Think. & 33,414 & 24,170 & 57,584 & 473 \\ 
Deepseek R1 & 15,242 & 28,021 & 43,263 & 458 \\
\midrule
\multicolumn{5}{l}{\textit{(Most Concise)}} \\
GPT-4o & 13,441 & 1,513 & 14,954 & 466 \\
\midrule
\textbf{Average} & 18,826 & 16,583 & 35,416 & 460 \\
\bottomrule
\end{tabular}
}
\end{table}

\vfill
\newpage

\section{Benchmark Composition}
\label{app:sec:detailed_metrics}

\begin{table}[h]
\centering
\small 
\caption{Biology Domain Composition. }
\label{tab:biology_composition}

\begin{tabularx}{\linewidth}{X r} 
\toprule
\textbf{Sub-Discipline} & \textbf{Count (\%)} \\
\midrule

\textbf{Molecular and Genomic Biology} & \textbf{29 (24.2\%)} \\
\hspace{1em} Genes, Proteins, and Nucleic Acids & 16 (13.3\%) \\
\hspace{1em} Gene Expression and Regulation & 7 (5.8\%) \\
\hspace{1em} DNA Structure and Dynamics & 5 (4.2\%) \\
\hspace{1em} Genetics and Inheritance & 1 (0.8\%) \\
\addlinespace 

\textbf{Ecology and Evolution} & \textbf{24 (20.0\%)} \\
\hspace{1em} Medical Biology and Pathology & 10 (8.3\%) \\
\hspace{1em} Evolutionary Biology and Genetics & 5 (4.2\%) \\
\hspace{1em} Biophysics and Theoretical Biology & 5 (4.2\%) \\
\hspace{1em} Ecology and Environmental Science & 4 (3.3\%) \\
\addlinespace

\textbf{Biotechnology and Bioengineering} & \textbf{20 (16.7\%)} \\
\hspace{1em} Molecular and Genetic Techniques & 9 (7.5\%) \\
\hspace{1em} Other Techniques and Models & 5 (4.2\%) \\
\hspace{1em} Imaging and Microscopy & 3 (2.5\%) \\
\hspace{1em} Drug Discovery and Delivery & 3 (2.5\%) \\
\addlinespace

\textbf{Systems and Omics Biology} & \textbf{12 (10.0\%)} \\
\hspace{1em} Systems-Level Concepts and Models & 8 (6.7\%) \\
\hspace{1em} "-omics" Fields & 4 (3.3\%) \\
\addlinespace

\textbf{Cellular and Subcellular Biology} & \textbf{10 (8.3\%)} \\
\hspace{1em} Cellular Processes & 3 (2.5\%) \\
\hspace{1em} Cell Death & 3 (2.5\%) \\
\hspace{1em} Organelles and Subcellular Structures & 2 (1.7\%) \\
\hspace{1em} Cytoskeleton and Molecular Motors & 2 (1.7\%) \\
\addlinespace

\textbf{Biochemistry and Metabolism} & \textbf{10 (8.3\%)} \\
\hspace{1em} Proteins and Enzymes & 8 (6.7\%) \\
\hspace{1em} Metabolic Pathways & 2 (1.7\%) \\
\addlinespace

\textbf{Immunology} & \textbf{9 (7.5\%)} \\
\hspace{1em} Core Concepts and Processes & 5 (4.2\%) \\
\hspace{1em} Signaling and Receptors & 4 (3.3\%) \\
\addlinespace

\textbf{Neuroscience} & \textbf{6 (5.0\%)} \\
\hspace{1em} Molecular and Cellular Neuroscience & 2 (1.7\%) \\
\hspace{1em} Systems and Cognitive Neuroscience & 2 (1.7\%) \\
\hspace{1em} Neuropharmacology & 1 (0.8\%) \\
\hspace{1em} Clinical Neuroscience & 1 (0.8\%) \\

\bottomrule
\end{tabularx}
\end{table}

\begin{table}[h]
\centering
\small 
\caption{Human Disease Domain Composition.}
\label{tab:human_disease_composition}

\begin{tabularx}{\linewidth}{X r} 
\toprule
\textbf{Sub-Discipline} & \textbf{Count (\%)} \\
\midrule

\textbf{Systemic \& Structural Genetic Disorders} & \textbf{35 (29.2\%)} \\
\hspace{1em} Multi-system Genetic Syndromes & 14 (11.7\%) \\
\hspace{1em} Skeletal and Overgrowth Syndromes & 13 (10.8\%) \\
\hspace{1em} Syndromes with Major Dermatologic Features & 4 (3.3\%) \\
\hspace{1em} Genomic and Inheritance Disorders & 4 (3.3\%) \\
\addlinespace

\textbf{Neurogenetic Disorders} & \textbf{18 (15.0\%)} \\
\hspace{1em} Neurodevelopmental Syndromes & 13 (10.8\%) \\
\hspace{1em} Hereditary Neuromuscular and Neurodegenerative Diseases & 5 (4.2\%) \\
\addlinespace

\textbf{Oncology} & \textbf{15 (12.5\%)} \\
\hspace{1em} Hematologic Malignancies & 6 (5.0\%) \\
\hspace{1em} Solid Tumors & 6 (5.0\%) \\
\hspace{1em} Hereditary Cancer Syndromes & 2 (1.7\%) \\
\hspace{1em} Transmissible Cancers & 1 (0.8\%) \\
\addlinespace

\textbf{Neurology} & \textbf{14 (11.7\%)} \\
\hspace{1em} Neurodegenerative and Movement Disorders & 4 (3.3\%) \\
\hspace{1em} Epilepsy and Seizure Disorders & 3 (2.5\%) \\
\hspace{1em} Other Neurological Conditions & 3 (2.5\%) \\
\hspace{1em} Neuroinflammatory and Demyelinating Diseases & 2 (1.7\%) \\
\hspace{1em} Congenital and Structural Brain Disorders & 2 (1.7\%) \\
\addlinespace

\textbf{Immuno \& Rheum} & \textbf{14 (11.7\%)} \\
\hspace{1em} Autoinflammatory and Autoimmune Syndromes & 7 (5.8\%) \\
\hspace{1em} Systemic Vasculitides & 4 (3.3\%) \\
\hspace{1em} Primary Immunodeficiencies & 2 (1.7\%) \\
\hspace{1em} Systemic and Catastrophic Conditions & 1 (0.8\%) \\
\addlinespace

\textbf{Endo \& Metab} & \textbf{13 (10.8\%)} \\
\hspace{1em} Inborn Errors of Metabolism & 11 (9.2\%) \\
\hspace{1em} Disorders of Mineral and Porphyrin Metabolism & 2 (1.7\%) \\
\addlinespace

\textbf{Organ-System} & \textbf{11 (9.2\%)} \\
\hspace{1em} Hematologic Disorders & 4 (3.3\%) \\
\hspace{1em} Gastroenterology and Hepatology & 3 (2.5\%) \\
\hspace{1em} Dermatologic Disorders & 1 (0.8\%) \\
\hspace{1em} Pulmonology & 1 (0.8\%) \\
\hspace{1em} Infectious and Transmissible Diseases & 1 (0.8\%) \\
\hspace{1em} Other Conditions & 1 (0.8\%) \\

\bottomrule
\end{tabularx}
\end{table}

\begin{table}[h]
\centering
\small 
\caption{Chemistry Domain Composition.}
\label{tab:chemistry_composition}

\begin{tabularx}{\linewidth}{X r} 
\toprule
\textbf{Sub-Discipline} & \textbf{Count (\%)} \\
\midrule

\textbf{Organic Chemistry} & \textbf{27 (22.5\%)} \\
\hspace{1em} Named Reactions and Rearrangements & 11 (9.2\%) \\
\hspace{1em} Synthesis, Reagents, and Specific Compounds & 8 (6.7\%) \\
\hspace{1em} Reaction Mechanisms and Concepts & 5 (4.2\%) \\
\hspace{1em} Stereochemistry and Molecular Structure & 3 (2.5\%) \\
\addlinespace

\textbf{Biochemistry and Chemical Biology} & \textbf{27 (22.5\%)} \\
\hspace{1em} Enzymes and Cofactors & 11 (9.2\%) \\
\hspace{1em} Biological Processes and Pathways & 8 (6.7\%) \\
\hspace{1em} Bio-analytical and Chemical Biology Tools & 6 (5.0\%) \\
\hspace{1em} Bio-inorganic Chemistry & 2 (1.7\%) \\
\addlinespace

\textbf{Physical Chemistry} & \textbf{19 (15.8\%)} \\
\hspace{1em} Quantum Chemistry and Spectroscopy & 6 (5.0\%) \\
\hspace{1em} Electrochemistry & 4 (3.3\%) \\
\hspace{1em} Thermodynamics and Calorimetry & 3 (2.5\%) \\
\hspace{1em} Kinetics and Dynamics & 3 (2.5\%) \\
\hspace{1em} Condensed Matter and Surface Chemistry & 3 (2.5\%) \\
\addlinespace

\textbf{Inorganometallics and Organometallics} & \textbf{17 (14.2\%)} \\
\hspace{1em} Organometallic Compounds & 8 (6.7\%) \\
\hspace{1em} Specific Inorganic Compounds & 6 (5.0\%) \\
\hspace{1em} Bonding and Molecular Structure & 3 (2.5\%) \\
\addlinespace

\textbf{Analytical Chemistry} & \textbf{11 (9.2\%)} \\
\hspace{1em} Spectroscopy and Spectrometry & 5 (4.2\%) \\
\hspace{1em} Instrumentation and Databases & 3 (2.5\%) \\
\hspace{1em} Microscopy and Surface Analysis & 2 (1.7\%) \\
\hspace{1em} Chromatography and Separation & 1 (0.8\%) \\
\addlinespace

\textbf{Applied Chemistry} & \textbf{8 (6.7\%)} \\
\hspace{1em} Photochemistry and Photophysics & 2 (1.7\%) \\
\hspace{1em} Magnetochemistry & 2 (1.7\%) \\
\hspace{1em} Industrial and Applied Chemistry & 2 (1.7\%) \\
\hspace{1em} Radiochemistry and Exotic Matter & 1 (0.8\%) \\
\hspace{1em} Theoretical Physics in Chemistry & 1 (0.8\%) \\
\addlinespace

\textbf{Materials Chemistry} & \textbf{6 (5.0\%)} \\
\hspace{1em} Key Concepts and Materials & 6 (5.0\%) \\
\addlinespace

\textbf{Theoretical Chemistry} & \textbf{5 (4.2\%)} \\
\hspace{1em} Concepts and Models & 3 (2.5\%) \\
\hspace{1em} Methods and Techniques & 2 (1.7\%) \\

\bottomrule
\end{tabularx}
\end{table}

\begin{table}[h]
\centering
\small 
\caption{Physics Domain Composition.}
\label{tab:physics_composition}

\begin{tabularx}{\linewidth}{X r} 
\toprule
\textbf{Sub-Discipline} & \textbf{Count (\%)} \\
\midrule

\textbf{Quantum Field Theory} & \textbf{22 (18.3\%)} \\
\hspace{1em} Foundational and Axiomatic QFT & 9 (7.5\%) \\
\hspace{1em} Gauge Theories and the Standard Model & 6 (5.0\%) \\
\hspace{1em} Non-Perturbative and Topological QFT & 3 (2.5\%) \\
\hspace{1em} Supersymmetry and Beyond & 2 (1.7\%) \\
\hspace{1em} QFT in Curved Spacetime & 2 (1.7\%) \\
\addlinespace

\textbf{Mathematical Physics} & \textbf{19 (15.8\%)} \\
\hspace{1em} Differential Geometry and Topology & 6 (5.0\%) \\
\hspace{1em} Group Theory and Algebra & 5 (4.2\%) \\
\hspace{1em} Functional Analysis and Operator Theory & 5 (4.2\%) \\
\hspace{1em} Integrable Systems and Exact Solutions & 2 (1.7\%) \\
\hspace{1em} Twistor Theory & 1 (0.8\%) \\
\addlinespace

\textbf{Quantum Mechanics} & \textbf{16 (13.3\%)} \\
\hspace{1em} Foundational Concepts and Formalism & 6 (5.0\%) \\
\hspace{1em} Quantum Systems and Dynamics & 4 (3.3\%) \\
\hspace{1em} Quantum Computation and Information & 4 (3.3\%) \\
\hspace{1em} Quantum States and Phenomena & 2 (1.7\%) \\
\addlinespace

\textbf{General Relativity} & \textbf{16 (13.3\%)} \\
\hspace{1em} Alternative and Extended Theories of Gravity & 7 (5.8\%) \\
\hspace{1em} Formalisms and Solutions & 5 (4.2\%) \\
\hspace{1em} Astrophysical Relativity & 3 (2.5\%) \\
\hspace{1em} Spacetime and Black Holes & 1 (0.8\%) \\
\addlinespace

\textbf{Specialized Topics in Physics} & \textbf{16 (13.3\%)} \\
\hspace{1em} Atomic, Molecular, and Optical (AMO) Physics & 7 (5.8\%) \\
\hspace{1em} Particle and Nuclear Physics & 5 (4.2\%) \\
\hspace{1em} Plasma Physics & 4 (3.3\%) \\
\addlinespace

\textbf{Condensed Matter} & \textbf{14 (11.7\%)} \\
\hspace{1em} Quantum Materials and Phases & 6 (5.0\%) \\
\hspace{1em} Electronic and Spintronic Effects & 4 (3.3\%) \\
\hspace{1em} Many-Body and Superconducting Systems & 2 (1.7\%) \\
\hspace{1em} Statistical and Soft Matter & 2 (1.7\%) \\
\addlinespace

\textbf{String Theory} & \textbf{12 (10.0\%)} \\
\hspace{1em} Core Concepts and Dualities & 6 (5.0\%) \\
\hspace{1em} Advanced and Related Theories & 5 (4.2\%) \\
\hspace{1em} Cosmological Applications & 1 (0.8\%) \\
\addlinespace

\textbf{Classical Physics and Electrodynamics} & \textbf{5 (4.2\%)} \\
\hspace{1em} Core Concepts and Phenomena & 5 (4.2\%) \\

\bottomrule
\end{tabularx}
\end{table}

\begin{table}[h]
\centering
\small 
\caption{Mathematics Domain Composition.}
\label{tab:mathematics_composition}

\begin{tabularx}{\linewidth}{X r} 
\toprule
\textbf{Sub-Discipline} & \textbf{Count (\%)} \\
\midrule

\textbf{Geometry \& Topology} & \textbf{37 (30.8\%)} \\
\hspace{1em} Algebraic Topology \& Homotopy Theory & 12 (10.0\%) \\
\hspace{1em} General \& Geometric Topology & 12 (10.0\%) \\
\hspace{1em} Differential \& Geometric Analysis & 11 (9.2\%) \\
\hspace{1em} K-Theory & 2 (1.7\%) \\
\addlinespace

\textbf{Number Theory} & \textbf{19 (15.8\%)} \\
\hspace{1em} Arithmetic \& Algebraic Geometry & 12 (10.0\%) \\
\hspace{1em} Algebraic Number Theory \& Class Field Theory & 5 (4.2\%) \\
\hspace{1em} Analytic Number Theory & 2 (1.7\%) \\
\addlinespace

\textbf{Analysis} & \textbf{18 (15.0\%)} \\
\hspace{1em} Operator Theory \& Operator Algebras & 9 (7.5\%) \\
\hspace{1em} Real, Complex, \& Harmonic Analysis & 8 (6.7\%) \\
\hspace{1em} Control Theory & 1 (0.8\%) \\
\addlinespace

\textbf{Cryptography \& Logic} & \textbf{18 (15.0\%)} \\
\hspace{1em} Cryptography & 7 (5.8\%) \\
\hspace{1em} Theoretical Computer Science \& Algorithms & 6 (5.0\%) \\
\hspace{1em} Logic \& Set Theory & 5 (4.2\%) \\
\addlinespace

\textbf{Algebra \& Group} & \textbf{14 (11.7\%)} \\
\hspace{1em} Homological \& Abstract Algebra & 8 (6.7\%) \\
\hspace{1em} Representation Theory & 4 (3.3\%) \\
\hspace{1em} Group Theory & 2 (1.7\%) \\
\addlinespace

\textbf{Theo Phys} & \textbf{8 (6.7\%)} \\
\hspace{1em} Quantum Mechanics \& Quantum Field Theory & 3 (2.5\%) \\
\hspace{1em} General Relativity \& Spacetime Physics & 2 (1.7\%) \\
\hspace{1em} Plasma \& Fluid Dynamics & 1 (0.8\%) \\
\hspace{1em} Physical Chemistry & 1 (0.8\%) \\
\hspace{1em} Dynamical Systems & 1 (0.8\%) \\
\addlinespace

\textbf{Statistics} & \textbf{6 (5.0\%)} \\
\hspace{1em} Probability \& Statistics & 4 (3.3\%) \\
\hspace{1em} Stochastic Analysis & 2 (1.7\%) \\

\bottomrule
\end{tabularx}
\end{table}

\begin{table}[h]
\centering
\small 
\caption{Computer Science Domain Composition.}
\label{tab:computer_science_composition}

\begin{tabularx}{\linewidth}{X r} 
\toprule
\textbf{Sub-Discipline} & \textbf{Count (\%)} \\
\midrule

\textbf{Theoretical Computer Science} & \textbf{43 (35.8\%)} \\
\hspace{1em} Mathematical Foundations & 25 (20.8\%) \\
\hspace{1em} Logic in Computer Science & 12 (10.0\%) \\
\hspace{1em} Computability and Complexity Theory & 6 (5.0\%) \\
\addlinespace

\textbf{Artificial Intelligence and Machine Learning} & \textbf{17 (14.2\%)} \\
\hspace{1em} Bayesian and Probabilistic Methods & 5 (4.2\%) \\
\hspace{1em} Learning Theory and Models & 4 (3.3\%) \\
\hspace{1em} Specialized Domains and Applications & 4 (3.3\%) \\
\hspace{1em} Logic, Reasoning, and Knowledge Representation & 3 (2.5\%) \\
\hspace{1em} Robotics and Embodied AI & 1 (0.8\%) \\
\addlinespace

\textbf{Programming Languages and Formal Methods} & \textbf{13 (10.8\%)} \\
\hspace{1em} Formal Verification & 4 (3.3\%) \\
\hspace{1em} Automata and Formal Languages & 4 (3.3\%) \\
\hspace{1em} Type Theory and Program Analysis & 3 (2.5\%) \\
\hspace{1em} Formal Semantics and Rewriting Systems & 2 (1.7\%) \\
\addlinespace

\textbf{Scientific Computing} & \textbf{13 (10.8\%)} \\
\hspace{1em} Computational Physics and Engineering & 6 (5.0\%) \\
\hspace{1em} Numerical Methods & 5 (4.2\%) \\
\hspace{1em} Optimization and Control & 2 (1.7\%) \\
\addlinespace

\textbf{Cryptography} & \textbf{12 (10.0\%)} \\
\hspace{1em} Cryptosystems and Primitives & 4 (3.3\%) \\
\hspace{1em} Cryptanalysis & 4 (3.3\%) \\
\hspace{1em} Secure Protocols & 2 (1.7\%) \\
\hspace{1em} Foundational Concepts & 2 (1.7\%) \\
\addlinespace

\textbf{Algorithmic Theory} & \textbf{11 (9.2\%)} \\
\hspace{1em} Graph and String Algorithms & 3 (2.5\%) \\
\hspace{1em} Algorithmic Information Theory & 3 (2.5\%) \\
\hspace{1em} Data Structures and Compression & 2 (1.7\%) \\
\hspace{1em} Combinatorial Optimization & 2 (1.7\%) \\
\hspace{1em} Algorithmic Game Theory & 1 (0.8\%) \\
\addlinespace

\textbf{Quantum Computing} & \textbf{11 (9.2\%)} \\
\hspace{1em} Quantum Algorithms and Information & 4 (3.3\%) \\
\hspace{1em} Quantum Foundations and Simulation & 4 (3.3\%) \\
\hspace{1em} Quantum Complexity and Models & 2 (1.7\%) \\
\hspace{1em} Interdisciplinary Applications & 1 (0.8\%) \\

\bottomrule
\end{tabularx}
\end{table}

\begin{table}[h]
\centering
\small 
\caption{Economics Domain Composition.}
\label{tab:economics_composition}

\begin{tabularx}{\linewidth}{X r} 
\toprule
\textbf{Sub-Discipline} & \textbf{Count (\%)} \\
\midrule

\textbf{Pluralist Econ} & \textbf{34 (28.3\%)} \\
\hspace{1em} Behavioral and Experimental Economics & 8 (6.7\%) \\
\hspace{1em} Public Economics and Social Choice Theory & 8 (6.7\%) \\
\hspace{1em} Marxian Economics & 7 (5.8\%) \\
\hspace{1em} Other Theories and Applied Fields & 6 (5.0\%) \\
\hspace{1em} Economic History and History of Thought & 5 (4.2\%) \\
\addlinespace

\textbf{Microeconomics} & \textbf{19 (15.8\%)} \\
\hspace{1em} Consumer and Firm Theory & 13 (10.8\%) \\
\hspace{1em} General Equilibrium and Market Design & 6 (5.0\%) \\
\addlinespace

\textbf{Econometrics} & \textbf{15 (12.5\%)} \\
\hspace{1em} Models and Methods & 11 (9.2\%) \\
\hspace{1em} Statistical Concepts & 4 (3.3\%) \\
\addlinespace

\textbf{Macroeconomics} & \textbf{14 (11.7\%)} \\
\hspace{1em} Economic Models and Growth & 8 (6.7\%) \\
\hspace{1em} Monetary Policy and Critiques & 6 (5.0\%) \\
\addlinespace

\textbf{Math/Comp Econ} & \textbf{14 (11.7\%)} \\
\hspace{1em} Models and Methods & 14 (11.7\%) \\
\addlinespace

\textbf{Game Theory} & \textbf{9 (7.5\%)} \\
\hspace{1em} Concepts and Models & 9 (7.5\%) \\
\addlinespace

\textbf{Finance} & \textbf{9 (7.5\%)} \\
\hspace{1em} Asset Pricing and Risk Management & 9 (7.5\%) \\
\addlinespace

\textbf{International Economics} & \textbf{6 (5.0\%)} \\
\hspace{1em} Trade and Finance & 6 (5.0\%) \\

\bottomrule
\end{tabularx}
\end{table}

\begin{table}[h]
\centering
\small 
\caption{Philosophy Domain Composition.}
\label{tab:philosophy_composition}

\begin{tabularx}{\linewidth}{X r} 
\toprule
\textbf{Sub-Discipline} & \textbf{Count (\%)} \\
\midrule

\textbf{Philosophy of Religion and Interdisciplinary} & \textbf{31 (25.8\%)} \\
\hspace{1em} History of Philosophy and Specific Texts & 10 (8.3\%) \\
\hspace{1em} Philosophy of Religion and Theology & 9 (7.5\%) \\
\hspace{1em} Other Interdisciplinary Concepts & 8 (6.7\%) \\
\hspace{1em} Philosophy of Language & 4 (3.3\%) \\
\addlinespace

\textbf{Logic and Philosophy of Mathematics} & \textbf{20 (16.7\%)} \\
\hspace{1em} Modal and Non-Classical Logics & 7 (5.8\%) \\
\hspace{1em} Logic in Computation and Language & 7 (5.8\%) \\
\hspace{1em} Set Theory and Mathematical Foundations & 6 (5.0\%) \\
\addlinespace

\textbf{Epistemology \& Phil Sci} & \textbf{18 (15.0\%)} \\
\hspace{1em} Epistemology & 10 (8.3\%) \\
\hspace{1em} Philosophy of Science & 4 (3.3\%) \\
\hspace{1em} Philosophy of Physics & 4 (3.3\%) \\
\addlinespace

\textbf{Metaphysics \& Ontology} & \textbf{16 (13.3\%)} \\
\hspace{1em} Core Concepts & 9 (7.5\%) \\
\hspace{1em} Theories of Reality & 4 (3.3\%) \\
\hspace{1em} Metaphysics of Mind & 3 (2.5\%) \\
\addlinespace

\textbf{Continental Philosophy} & \textbf{13 (10.8\%)} \\
\hspace{1em} Phenomenology and Existentialism & 5 (4.2\%) \\
\hspace{1em} Post-Structuralism and Deconstruction & 4 (3.3\%) \\
\hspace{1em} Lacanian Psychoanalysis & 4 (3.3\%) \\
\addlinespace

\textbf{Eastern Philosophy} & \textbf{12 (10.0\%)} \\
\hspace{1em} Indian Philosophy (Hindu Schools) & 6 (5.0\%) \\
\hspace{1em} Indian Philosophy (Buddhist Schools) & 5 (4.2\%) \\
\hspace{1em} Indian Religious/Philosophical Concepts & 1 (0.8\%) \\
\addlinespace

\textbf{Social Philosophy} & \textbf{6 (5.0\%)} \\
\hspace{1em} Legal and Social Philosophy & 5 (4.2\%) \\
\hspace{1em} Political Philosophy & 1 (0.8\%) \\
\addlinespace

\textbf{Mind} & \textbf{4 (3.3\%)} \\
\hspace{1em} Core Concepts & 3 (2.5\%) \\
\hspace{1em} Psychoanalytic Philosophy & 1 (0.8\%) \\

\bottomrule
\end{tabularx}
\end{table}

\clearpage
\section{Active Probing Prompt}
\label{app:sec:prompts}

\begin{strip} 
\centering
    
\begin{mytextbox}[Querying Agent Prompt]
You are a \texttt{\{domain\}} expert. I will have a preset \texttt{\{domain\}} name/term/concept in mind, and you need to continuously ask questions and reason until you can identify the name/term/concept. You can ask questions to narrow down the possibilities. Your questions must be in the form of a single, four-option multiple-choice question at a time.
\\
You can only ask one multiple-choice question at a time. Each question is crucial and should be based on your reasoning, professional knowledge, and the previous rounds of questions and answers. Please try to identify the term/concept in as few rounds as possible.
\\
I have now preset a name/term/concept. Please begin with your first question.
\end{mytextbox}
\captionof{figure}{The system prompt is for the querying agent. In each round, it poses strategic queries to quickly identify the $P_{\text{target}}$ by interacting with an environment. The \texttt{\{domain\}} placeholder is replaced with the specific domain name (e.g., Human Disease, Physics) for each evaluation.}
\label{fig:system_prompt}

\vspace{1em} 
    
\begin{mytextbox}[Knowledge Oracle Prompt]
You are a \texttt{\{domain\}} expert. You will have a preset \texttt{\{domain\}} name/term/concept in mind, and I need to identify it by asking questions and using reasoning. I can ask questions to narrow down the scope, and my method of questioning will be a single, four-option multiple-choice question at a time. The specific details regarding my question are all within the \texttt{<qa\_response>} part. You will respond to my questions based on some ``golden information'' (reference material).
\\
The name/term/concept you have preset is: \texttt{\{puzzle\_name\}} \\
Reference material: \texttt{<reference\_material>} \texttt{\{puzzle\_reference\}} \texttt{</reference\_material>}
\\
In your reply, you must include a \texttt{<pure\_parse>...</pure\_parse>} block and a \texttt{<pure\_answer>...</pure\_answer>} block.
\\
\textbf{For the \texttt{<pure\_parse>...</pure\_parse>} block:} \\
You will only judge the correctness of the options. For example, when option A, D are correct, and option B, C are incorrect: \\
\texttt{<pure\_parse>A is correct, B is incorrect, C is incorrect, D is correct</pure\_parse>} \\
When all options are incorrect, reply with \texttt{<pure\_parse>A is incorrect, B is incorrect, C is incorrect, D is incorrect</pure\_parse>} \\
If the \texttt{<qa\_response>} part provides only an incorrect final guess without including a multiple-choice question, reply with \texttt{<pure\_parse>Your guess is incorrect, please ask a new question</pure\_parse>} and \texttt{<pure\_answer>None</pure\_answer>}. \\
The \texttt{<pure\_parse>} block is forbidden from containing the name of the preset name/term/concept or any other content from the \texttt{<reference\_material>} part.
\\
\textbf{For the \texttt{<pure\_answer>...</pure\_answer>} block:} \\
If the message you receive in the \texttt{<qa\_response>} part (or one of the options in the \texttt{<qa\_response>} part) explicitly states the name of your preset name/term/concept (or its alias), then include \texttt{<pure\_answer>Your preset name</pure\_answer>} in your reply (Here, ``Your preset name'' is replaced with the actual name that appears). If it does not, then include \texttt{<pure\_answer>None</pure\_answer>}.
\\
All of your replies must include both of these blocks.
\end{mytextbox}

\captionof{figure}{The system prompt for the Knowledge Oracle. This agent acts as an oracle, providing structured and non-leaking feedback to the strategic agent based on the $P_{\text{target}}$ and $I_\text{ref}$.}
\label{fig:environment_prompt}

\end{strip}

\end{document}